\documentclass[letterpaper, 10 pt, conference]{ieeeconf}  

\IEEEoverridecommandlockouts                              

\usepackage{pst-grad}
\usepackage{multido} 
\usepackage{calc}
\usepackage{fp-snap} 
\usepackage{ifthen}
\usepackage{graphicx}
\usepackage{float}
\usepackage{epsfig} 
\usepackage{mathptmx} 
\usepackage{times} 
\usepackage{amsmath} 
\usepackage{amssymb}  
\usepackage[utf8]{inputenc}
\usepackage{upgreek} 
\usepackage{cancel} 
\usepackage{mathdots} 
\usepackage{mathrsfs} 
\usepackage{stackrel} 
\usepackage[colorlinks=true,linkcolor=magenta]{hyperref} 
\hypersetup{
    colorlinks=true,
    linkcolor=blue,
    filecolor=magenta, 
    citecolor=blue,
    urlcolor=blue,
    pdftitle={How to build and validate a safe and reliable Autonomous Driving stack? A ROS based software modular architecture baseline},
    pdfpagemode=FullScreen,
    }
\usepackage{algorithm,algorithmic} 
\usepackage{multirow} 
\usepackage{subfigure}

\usepackage{verbatim}
\newcommand{\keywordsnew}[1]{\textbf{\textit{Keywords: }}{#1}}
\usepackage{color, colortbl}
\definecolor{ForestGreen}{RGB}{34,139,34}
\usepackage{amsmath}

\usepackage{graphics} 
\usepackage{mathptmx} 
\usepackage{amsmath} 
\usepackage{graphicx}
\usepackage{array}
\usepackage{tabu} 
\usepackage{lipsum}
\usepackage{multicol}
\usepackage{multirow}
\usepackage{mathtools}
\usepackage{commath}
\usepackage{verbatim} 
\usepackage{cleveref}
\usepackage{array}
\usepackage{diagbox}
\usepackage{textcomp}

\newcommand\colouredcite[1]{\textcolor{green}{\cite{#1}}}

\usepackage[
  separate-uncertainty = true,
  multi-part-units = repeat
]{siunitx}

\usepackage{graphicx}
\usepackage{xcolor}
\usepackage{blindtext}

\title{\LARGE \bf
Exploring Attention GAN for Vehicle Motion Prediction
}

\author{Carlos Gómez-Huélamo$^{1}$, Marcos~V.~Conde$^{2}$, Miguel Ortiz$^{1}$, \\
Santiago Montiel$^{1}$, Rafael Barea$^{1}$ and Luis M. Bergasa$^{1}$
\thanks{$^{1}$Carlos Gómez-Huélamo, Miguel Ortiz, Santiago Montiel, Rafael Barea and Luis M. Bergasa are with the Department of Electronics, University of Alcal{\'a} (UAH), Spain. {{carlos.gomezh, eduardo.ortiz, santiago.montiel, rafael.barea, luism.bergasa\}@uah.es}}, $^{2}$Marcos Conde is with the University of Würzburg, Computer Vision Lab, CAIDAS.}
}

\begin{document}

\maketitle
\thispagestyle{empty}
\pagestyle{empty}

\begin{abstract}

The design of a safe and reliable Autonomous Driving stack (ADS) is one of the most challenging tasks of our era. These ADS are expected to be driven in highly dynamic environments with full autonomy, and a reliability greater than human beings. In that sense, to efficiently and safely navigate through arbitrarily complex traffic scenarios, ADS must have the ability to forecast the future trajectories of surrounding actors. Current state-of-the-art models are typically based on Recurrent, Graph and Convolutional networks, achieving noticeable results in the context of vehicle prediction. In this paper we explore the influence of attention in generative models for motion prediction, considering both physical and social context to compute the most plausible trajectories. We first encode the past trajectories using a LSTM network, which serves as input to a Multi-Head Self-Attention module that computes the social context. On the other hand, we formulate a weighted interpolation to calculate the velocity and orientation in the last observation frame in order to calculate acceptable target points, extracted from the driveable of the HDMap information, which represents our physical context. Finally, the input of our generator is a white noise vector sampled from a multivariate normal distribution while the social and physical context are its conditions, in order to predict plausible trajectories. We validate our method using the Argoverse Motion Forecasting Benchmark 1.1, achieving competitive unimodal results. Our code is publicly available at \href{https://github.com/Cram3r95/mapfe4mp}{\fbox{https://github.com/Cram3r95/mapfe4mp}}.

\end{abstract}

\keywordsnew{Autonomous Driving, Motion Prediction, Generative Adversarial Networks, Self-Attention, LSTM}

\section{INTRODUCTION}
\label{section:introduction}

\begin{figure}[!ht]
  \centering
   \includegraphics[width=0.85\linewidth]{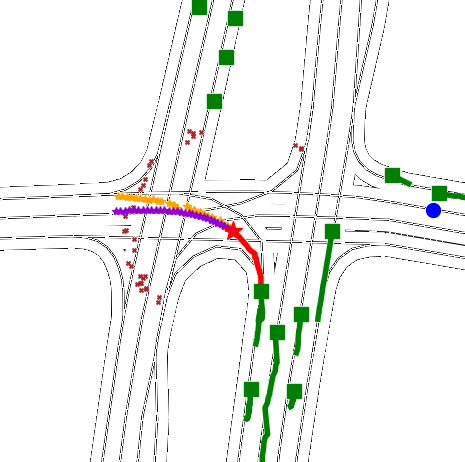}
   \caption{MP Scenario in Argoverse~\colouredcite{chang2019argoverse}. We represent: our vehicle (\textcolor{blue}{ego}), the target \textcolor{red}{agent}, and \textcolor{ForestGreen}{other agents}. We also show the \textcolor{orange}{groundtruth} trajectory, the \textcolor{purple}{prediction} and potential \textcolor{brown}{goal-points}. Markers are current positions.}
   \label{fig:teaser}
\end{figure}

In order to achieve a reliable navigation, Autonomous Driving stacks (ADS) have to perform safe driving behaviours following conventional traffic rules. Regarding this, one of the most challenging tasks of an ADS is to forecast the future motion of surrounding actors, used by AD algorithms such as target selection and path planning throughout the navigation. To capture the complexity of an arbitrarily complex scenario, a robust and reliable motion prediction model should not only take into account the past trajectory of the most important agents around the vehicle, but also physical information (such as on-board sensor information or prior knowledge identified with map information \colouredcite{murciego2018topological}) and the interaction of the ego-vehicle with the environment. 

\begin{figure*}[!ht]
	\centering
	\includegraphics[width=\textwidth]{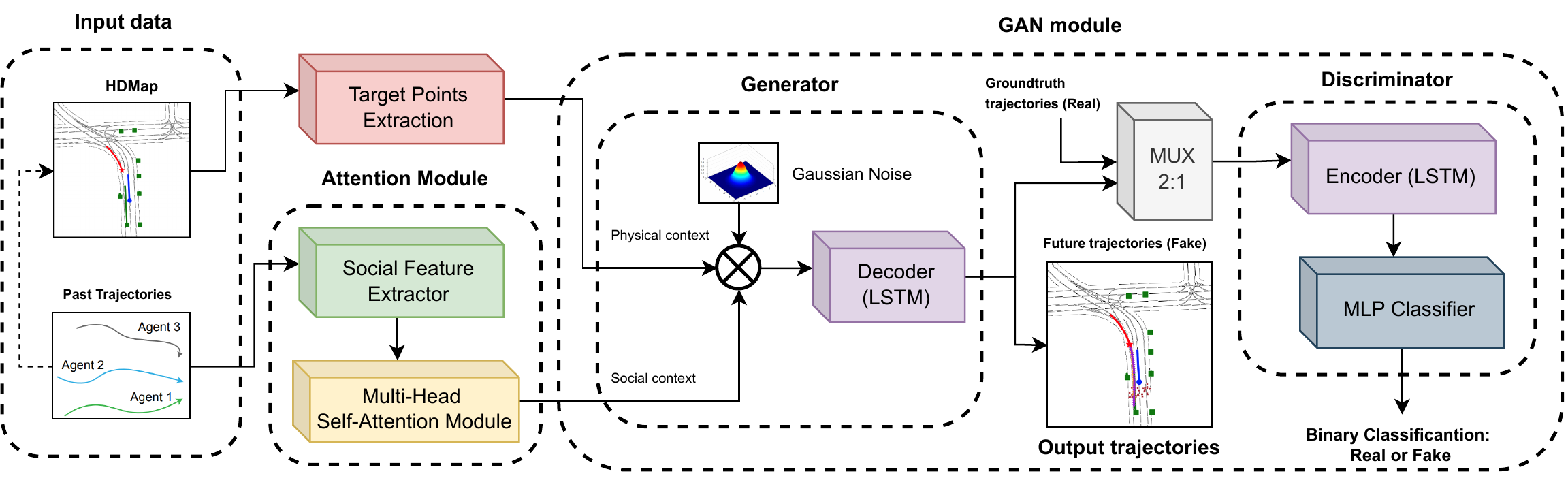}
	\caption{Overview of our \textbf{Attention-based Generative Model}. We distinguish three main blocks: 1) \textbf{Target points extractor}, which uses HDMap information and agent past trajectory to compute acceptable target points~\colouredcite{gomez2022exploring}, 2) \textbf{Attention module}, responsible for encoding the trajectories of the surrounding vehicles and applying Multi-Head Self-Attention, 3) \textbf{LSTM based GAN module}, which consists of a LSTM based decoder as the ''generator", in charge of taking into account the estimated target locations and the dynamic feature to generate the future trajectories, and a discriminator to force the generator to produce more realistic predictions.}
	\label{fig:general_architecture_system_pipeline}
	\vspace{-20pt}
\end{figure*}

The main challenge in this task is the human driver behaviour can neither be modeled and consequently predicted properly, specially in negotiating situations \colouredcite{gomez2021train} \colouredcite{mercat2020multiattentmotion} with many participants where considering agent-environment/agent-agent interactions \colouredcite{sadeghian2019sophie} plays a determinant role. Then, resulting trajectories may not be necessarily feasible, not covering the full spectrum of possible trajectories that a vehicle can take. In that sense, a more natural way of capturing the feasible directions \colouredcite{dendorfer2020goalgan} is to first compute a set of intermediate target points from a distribution of acceptable positions. In this work we explore the influence of attention mechanisms in generative models, in particular based on Generative Adversarial Network (GAN) \colouredcite{goodfellow2014generative}, to carry out the task of motion prediction. Our model considers both physical context, computing acceptable target points from the driveable area around the target agent, and social context, LSTM (Long Short-Term Memory) \colouredcite{hochreiter1997long} based encoder as input to a Multi-head self-attention module, as input of our generator, which combines the scene understanding around the agent vehicle (target agent to predict its trajectory) and the corresponding noise vector associated to generative models to compute the trajectories using a LSTM decoder, as illustrated in Fig. \ref{fig:general_architecture_system_pipeline}. In this context, the discriminator is applied in order to force the generator model to produce more realistic samples (trajectories, as shown in Fig. \ref{fig:teaser}), hence, to improve the performance. 

The remaining content of this work is organized as follows. The next section reviews some state-of-the-art algorithms for vehicle motion prediction. Section~\ref{section:our_approach} presents our pipeline, focused on a combination of precomputed target points, Multi-Head~\colouredcite{zhang2019multi} Self Attention and LSTM-based Motion Encoder to generate the most plausible trajectories in a Routing Module (LSTM based decoder), framed by a generative adversarial training to model the stochastic nature of the process. Section~\ref{section:experimental_results} presents quantitative and qualitative results in the Argoverse Benchmark, illustrating the strengths and weaknesses of our model. Finally, Section~\ref{section:conclusions_and_future_works} concludes the paper.

\section{RELATED WORKS}
\label{section:related_works}

One of the crucial tasks that an ADS must face during navigation, specially in arbitrarily complex urban scenarios, is to predict the behaviour for moving objects. In this section we review recent literature on motion prediction, illustrating different works focused on pedestrian prediction and vehicle prediction, as well as social attention to capture complex social interactions among agents. 

\subsection{Motion Prediction}

Prior knowledge on motion prediction in pedestrian datasets \colouredcite{pellegrini2009ethdata} \colouredcite{lerner2007ucydata} usually focuses on Deep Learning methods such as Long Short-Term Memory (LSTM)~\colouredcite{hochreiter1997long} and Generative Adversarial Network (GAN)~\colouredcite{goodfellow2014gan}. SocialLSTM \colouredcite{alahi2016social} proposes an LSTM model that can jointly predict the paths of all agents in the scene taking into account the common sense rules and social conventions using a social-pooling module. SocialGAN \colouredcite{gupta2018social} enhances SocialLSTM with a generative adversarial framework, introducing a variety loss that encourage the network to cover the space of plausible paths and proposing a novel pooling global social pooling vector that encodes the subtle cues for all agents involved in the scene. SoPhie \colouredcite{sadeghian2019sophie} considers not only the path history of all agents but also the physical context information (captured by a top-view static image, computing salient regions of the scene), combining physical and social attention mechanisms in order to help the model knows what to extract and where to focus. Goal-GAN \colouredcite{dendorfer2020goalgan} predicts the most likely goal points of the agent of interesting, estimating a set of trajectories towards these goals using both physical and social context, as proposed by \colouredcite{sadeghian2019sophie}. On the other hand, in the context of vehicle prediction \colouredcite{chang2019argoverse} \colouredcite{caesar2020nuscenes}, prior information takes more importance regarding the risk at certain velocities in urban/highway environments in order to perform safe navigation. High-fidelity maps (hereafter, HDMaps) have been widely adopted to provide offline (also known as context) information to complement the online information provided by the sensor suite of the vehicle and its corresponding algorithms. Recent learning-based approaches \colouredcite{hong2019rules}\colouredcite{dong2021multimodal}\colouredcite{chai2019multipath} \colouredcite{gao2020vectornet} present the benefit of having probabilistic interpretations of different behaviour hypotheses, requiring to build a representation to encode the trajectory and map information.   

\subsection{Social Attention}

In a similar way to humans that pay more attention to close obstacles, people walking towards them or upcoming turns rather than considering the presence of people or building far away, the perception layer of a self-driving car must be modelled to focus more on the more relevant features of the scene. Social Attention is a mechanism that allows selective interactions within relevant agents. SoPhie \colouredcite{sadeghian2019sophie} computes a different context vector for each agent, in such a way other agents features are sorted in terms of their relative distance to the agent of interest. Then, a soft attention mechanism is used to compute a context feature vector, which represents the social context. Nevertheless, a fixed size ($N_{\text{max}}$ agents) list that considers the context of all agents is sensitive to small variations \colouredcite{mercat2020multiattentmotion} of other agents positions. In that sense, SocialWays \colouredcite{amirian2019social} presents a hand-crafted relative geometric feature to produce a set of normalized weights, in such a way the context vector represents a convex sum of other feature vectors (context of each agent) that is invariant to the ordering. Nevertheless, these attention mechanisms were not designed to model complex interactions, no more than angles and distances due to the inherent problem of pedestrian prediction, in such a way we must find this challenging interactions in the vehicle motion prediction task to account for specific behaviours like overtaking, Adaptive Cruise Control (ACC), emergency braking or yielding. GRIP \colouredcite{li2019grip} proposes a graph representation of vehicle neighbours, taking into account local interactions among vehicles withing a \textit{d} distance threshold around the target agent. 

Vemula et~al.~\colouredcite{vemula2018social} uses a dot product attention module (inspired from the attention mechanism proposed by \colouredcite{vaswani2017attention} for sentence translation), allowing joint forecast of every agent in the scene without spatial limitations, considering long range interactions regardless the ordering of the input vehicles tracks and the number of vehicles. Moreover, \colouredcite{vemula2018social} combines this dot product with a spatio-temporal graph representation to take into account temporal and spatial dependencies of the agents, such as their absolute/relative positions and time step movements. Mercat et~al.~\colouredcite{mercat2020multiattentmotion} presents a multi-head extension of this dot attention mechanism, where each agent is embedded by means LSTMs before computing the dot product attention in order to produce social interactions. 

\section{OUR APPROACH}
\label{section:our_approach}

In this work we aim to develop a model that can successfully predict plausible future trajectories in the context of vehicle prediction, taking into account not only the past trajectory of the corresponding agent but also the past state of the most relevant obstacles around it and HDmap information to compute a set of acceptable target points representing the physical constraints for our problem.

\subsection{Problem Definition}

We tackle the task of predicting the future positions of certain agents. Each position throughtout the whole sequence (past observations and future observations) is expressed via via $x$ and $y$ coordinates in the 2D ground plane. As input, we are given the past trajectory of all agents in the scene as well as prior knowledge, identified with HDMap information in the case of Argoverse benchmark 1.1. We observe the trajectories $X_i = \{ \left(x^t_i, y^t_i \right)\in \mathbb{R}^2 \vert t\,=\,1, \dots, t_{obs}$ \} of $N$ agents in the scene and the corresponding physical information of the scene (2D HDMap), observed at the timestep $t_{obs}$. Our goal is to predict the future positions $Y_i = \{ \left(x^t_i, y^t_i \right) \in \mathbb{R}^2 \vert t\,=\,t_{obs}+ 1, \dots, t_{pred}\}$ of a particular agent, also referred as the target agent. These future trajectories should be compliant with the social (i.e. traffic rules, such as right-a-way, crosswalk, left/right turning) and physical (ensuring the presence of the vehicle in the driveable area) constraints of the scene.

\subsection{Overall Model}

When vehicles drive through a traffic scenario, they usually aim to reach partial goals, depending on their predefined navigation route and scene context (both physical and social), until they finally arrive at their final destination. Formally, given a certain goal, vehicles must face different traffic rules and other agents along their way to reach their final destination. Regarding this, our model takes computes both the social context and acceptable target points for the corresponding agent given its past trajectory and then generates plausible trajectories towards the estimated goals. Our model consists of three main blocks:

\begin{itemize}
    \item \textit{Target Points Extractor}: Combines HDMap information and dynamic features of the target agent (speed and orientation) to generate acceptable target points in the driveable area.
    \item \textit{Attention module}: Computes the agent's dynamic features recursively by means a LSTM unit and uses a Multi-Head Self Attention~\colouredcite{zhang2019multi, mercat2020multiattentmotion} module to capture complex social interactions among agents.
    \item \textit{GAN module}: Given the target points and highlighted social features, this module generates plausible and realistic trajectories using a LSTM based decoder, which represents the generator. Discriminator is applied to enhance the performance of the generator by forcing it to compute more realistic predictions.
\end{itemize} 

Fig. \ref{fig:general_architecture_system_pipeline} illustrates an overview of our model. Next, we describe the different blocks of our model.

\subsubsection{Target points extraction}

In order to compute acceptable destination, also referred as target points, and integrate them into the model as a prior, we make use of the HDMap information around the target agent and a simple yet powerful method proposed by \colouredcite{tang2021exploring} to compute the velocity and orientation of the target agent in the last observation frame. First, we calculate the driveable area around the vehicle considering a hand-defined \textit{d} threshold. Then, we consider the dynamic features of the target agent in the last observation frame $t_{obs}$ to compute acceptable target points in local coordinates. After estimating these variables, given the driveable area around the target agent, we randomly $L$ target points considering a constant velocity model during the prediction horizon \textit{$t_{pred}$} and the estimated orientation, assuming non-holonomic constraints which are inherent of standard road vehicles, that is, the car has three degrees of freedom, its position in two axes and its orientation, and must follow a smooth trajectory in a short-mid term prediction. For further details, we refer the reader to \colouredcite{gomez2022exploring}.

\subsubsection{Attention module}

Our model takes as social input the past $n$ observations in map (global) coordinates for each agent in the scene, encoding these trajectories as a preliminary stage before feeding a Multi-Head Self-Attention (MHSA)~\colouredcite{vaswani2017attention} module that computes the social context of the scene, as observed in Fig. \ref{fig:general_architecture_system_pipeline}. The past trajectory of an agent is transformed to relative (local coordinates) displacement vectors $\left( \Delta x_i^t, \Delta y_i^t \right)$ and embedded into a higher dimensional vector with a Multi-Layer Perceptron (MLP), which serves as input of the LSTM unit, dynamic feature extractor to capture the speed and direction of the corresponding agent. Then, the hidden state of the LSTM ($h_{M\!E}$) is used by the MHSA module that learns complex social interactions while being invariant to their number and ordering, avoiding a fixed size ($N_{\text{max}}$ agents) list which would be sensitive to small variants in the agent's positions. In this context, each agent of the scene should pay attention to specific features from the most relevant agents around it. The Multi-Head Self-Attention module consists of several heads that given the encoded trajectories produces feature vectors that encode all pairwise relations among agent's information. Implementation details are specified in Section~\ref{section:experimental_results}.

\subsubsection{GAN module}

As stated in Section \ref{section:related_works}, to capture the stochastic nature of motion prediction, state-of-the-art methods leverage the power of generative models, such as Variational Autoencoders (VAEs) and Generative Adversarial Networks (GANs). In our work we use an adversarial framework in order to train our trajectory generator, responsible for generating physically and realistic feasible trajectories. In a GAN, the Generator (which after being trained will be the inference network) and Discriminator networks compete in a two-player min-max game \colouredcite{goodfellow2014gan}, as observed in Eq. \ref{eq:min_max_game}. While the generator aims at producing feasible trajectories, the discriminator learns to differentiate between fake and real samples, in other words, groundtruth (which are feasible by definition) and inferred trajectories, in such a way the tasks of the discriminator is to enhance the performance of the generator by forcing it to compute more realistic predictions, more and more similar to the groundtruth trajectory. As a result, the generator should be able to produce outputs which the discriminator cannot discriminate clearly, indicating that the output is realistic. 

\begin{eqnarray}
\label{eq:min_max_game}
&&\hspace{-10mm} \min_{Gen} \max_{Dis} V(Dis, Gen)=E_{X \sim p_{data}(X)}[\mbox{log} Dis(X,Y)] \nonumber \\
&&\hspace{15mm} + E_{z \sim p_z(z)}[\mbox{log} (1 - Dis(X, Gen(X,z)))],
\end{eqnarray}

In the present case, the generator, also identified as the routing module, is represented by a decoder LSTM  ($LSTM_{gen}$) and the discriminator by a classifier LSTM ($LSTM_{dis}$) so as to estimate the temporally dependent future states. Similar to the conditional GAN proposed by \colouredcite{sadeghian2019sophie}, the input to our generator is a concatenation of a white noise vector $z$ sampled from a multivariate normal distribution, being the physical context (goal points in relative coordinates in the last observation frame, $C_{Ph(i)}^{t_{obs}}$) and social context (interactions among agents, $C_{So(i)}^{1:t_{obs}}$) its conditions. Then, the generated future trajectory for a particular agent is modelled as Eq. \ref{eq:gen_dec}:

\begin{eqnarray}
\label{eq:gen_dec}
& \hat{Y}_i^{t_{obs}:t_{pred}} = LSTM_{gen}\big(C_{Ph(i)}^{t_{obs}}; C_{So(i)}^{1:t_{obs}}; z\big)
\end{eqnarray}

On the other hand, the input of the discriminator is a randomly chosen trajectory sample from the either predicted future trajectory or groundtruth for the corresponding agent up to $t = t_{obs} + t_{pred}$ frame, i.e.  $T_i^{t_{obs}:t_{pred}}\sim p(\hat{Y}_i^{t_{obs}:t_{pred}},Y_i^{t_{obs}:t_{pred}})$.

\begin{eqnarray}
\label{eq:dis}
\hat{L}_{i}^{t_{obs}:t_{pred}} = LSTM_{dis}(T_i^{t_{obs}:t_{pred}})
\end{eqnarray}

Then, the discriminator returns a label $\hat{L}_{i}^{t_{obs}:t_{pred}}$ for the chosen trajectory indicating whether the trajectory is groundtruth (real) $Y_i^{t_{obs}:t_{pred}}$ or predicted (fake) $\hat{Y}_i^{t_{obs}:t_{pred}}$, being the labels 0 and 1 for fake and real trajectories respectively. Eq. \ref{eq:dis} summarizes the discriminator working principles. 

\subsubsection{Losses}

To train our model, we use the following losses:

\begin{eqnarray}
\label{eq:obj}
W^* =\operatorname*{argmin}_W \quad\mathbb{E}_{i,\tau}[\lambda_{gan} \mathcal{L}_{GAN}\big(\hat{L}_{i}^{t_{obs}:t_{pred}}, L_{i}^{t_{obs}:t_{pred}} \big)+ \nonumber\\
\lambda_{ade} \mathcal{L}_{L2}(\hat{Y}_i^{t_{obs}:t_{pred}},Y_i^{t_{obs}:t_{pred}})+ \nonumber\\
\lambda_{fde} \mathcal{L}_{L2}(\hat{Y}_i^{t_{obs}+t_{pred}},Y_i^{t_{obs}+t_{pred}})],
\end{eqnarray}

where $W$ is the collection of the weights of all networks used in our model and the different $\lambda$ represent the corresponding regularizers between these losses. As stated in Eq. \ref{eq:min_max_game}, $\mathcal{L}_{GAN}$ represents the min-max game where the generator tries to minimize the function while the discriminator tries to maximize it. ADE loss function is commonly used to compute the average error between the predicted trajectories and the corresponding groundtruth. Moreover, we add FDE loss function to explicitly optimize the distribution towards the final real point.   

\section{EXPERIMENTAL RESULTS}
\label{section:experimental_results}

\subsection{Dataset}

We evaluate our work on the well-established and public available Argoverse Motion Forecasting dataset \colouredcite{chang2019argoverse}, including the training, validation and testing subsets from its official website~\colouredcite{argobench}. 

It consists of 205942 training samples, 39472 validation samples and 78143 samples. Each sample in the Argoverse Dataset has a length of 5 seconds, with an observation window of 2 seconds and a prediction window of 3 seconds, including the corresponding labels of the agents (AGENT, as the target agent, AV, the vehicle that captures the scene and OTHER, representing the remaining relevant obstacles) and a global map from the cities of Pittsburgh and Miami. The sampling frequency is $10\mathrm{Hz}$. The main goal here is to predict the 3s future position of the target agent in the scene, which is supposed to be the vehicle that faces the most challenging traffic scenarios.

\subsection{Metrics}

Previous works \colouredcite{chai2019multipath, mercat2020multiattentmotion, sadeghian2019sophie} report the minimum Average Displacement Error ($\text{minADE}_K$), which averages the $L2$ distances between the ground truth and predicted output across all timesteps and minimum Final Displacement error ($\text{FDE}_K$), which computes the $L2$ distance between the final points of the ground-truth and the predicted final position, taking the best $K$ trajectory sample of each agent compared to the ground truth. In the present work, we use $K$ = 1 (unimodal case).
\\

\subsection{Implementation details}

All local test were conducted in a PC desktop (AMD Ryzen 9 5900X, 32GB RAM with CUDA-based NVIDIA GeForce RTX 3090 24GB VRAM, Ubuntu 18.04).

We design our dataloader to sample in each batch a 30/70 proportion of straight and curved trajectories (regarding the target agent's whole trajectory). We classify a trajectory as straight or curve estimating a first degree trajectory by means the RANSAC algorithm with the highest number of inliers (tolerance $t$ set to 2m, max trials=30, min samples=60\% total observations). Then, if the actual trajectory presents 20\% or more consecutive points further than $t$ with respect to the closest point of the fitted trajectory, the whole sequence is labelled as curve. We do this to focus in the training process in non-linear prediction, which represents one the key challenges in vehicle motion prediction. 

Regarding the ablation study, we train the different models for 150 epochs using Adam optimizer with learning rate $0.001$ and default parameters, linear LR Scheduler with factor $0.5$ decay on plateaus (5k iterations) and batch size $64$. The loss function is weighted by setting $\lambda_{gan}$=1.4, $\lambda_{ade}$=1 and $\lambda_{fde}$=1.5, giving more importance to the adversarial loss and the final displacement error. Similar to \colouredcite{sadeghian2019sophie}, the LSTM encoder (attention block) encodes trajectories using a single layer MLP with an embedding dimension of 16. We set all LSTM units to have $32$ hidden dimensions. The number of target points is set also to 32 in order to compute the physical context. Moreover, in order to calculate these target points we consider the same prediction horizon $t_{pred}=3s$ to estimate the distance travelled assuming a constant velocity model. To make our model more robust to scene orientation, we augment the training data adding some white noise ($\mu=0, \sigma=0.25$, [m]) to the observation data, rotating the scene and also dropping and replacing (with their last frame) some observations of the past trajectory in order to make the trained model general enough so as to perform well on the unseen traffic scenarios in the split test which different scene geometries such as left/right turning or emergency braking.

\subsection{Model results}

In this section, we perform an ablation study and compare our method's performance against state-of-the-art results on the Argoverse Motion Forecasting benchmark (split test). Additionally, we conduct a statistical analysis on the Argoverse validation set for the ADE and FDE metrics, distinguishing the performance between straight and curved trajectories.

\begin{table}[!h]
\caption{Ablation study of our unimodal pipeline, and comparison with other relevant methods on Argoverse. We can see the improvement using Target points (TP) and Class balance (CB)}
    \begin{center}
        \begin{tabular}{| l | c | c |}
            \hline
            \textbf{Model} & \textbf{ADE (k=1) $\downarrow$} & \textbf{FDE (k=1) $\downarrow$} \\
            & [m] & [m] \\
            \hline
            Constant Velocity~\colouredcite{chang2019argoverse} & 3.53 & 7.89 \\ 
            Argoverse Baseline (NN)~\colouredcite{chang2019argoverse} & 3.45  & 7.88 \\ 
            Argoverse Baseline (LSTM)~\colouredcite{chang2019argoverse} & 2.96  & 6.81 \\ 
            SGAN~\colouredcite{gupta2018sgan} & 3.61  & 5.39 \\ 
            TPNet~\colouredcite{fang2020tpnet} & 2.33  & 5.29 \\ 
            TPNet-map~\colouredcite{fang2020tpnet} & 2.33  & 4.71 \\ 
            Jean (1st)~\colouredcite{chang2019argoverse, mercat2020multiattentmotion} & 1.74  & 4.24 \\ 
            \hline
            Ours Baseline (*) & 1.98  & 4.47 \\ 
            Ours + TP & 1.78  & 4.13 \\
            Ours + CB & 1.82  & 4.09 \\
            Ours + TP + CB & 1.67  & 3.82 \\
            \hline
        \end{tabular}
        \label{tab:test_comparison}
    \end{center}
    \vspace{-10pt}
\end{table}

Table \ref{tab:test_comparison} illustrates the comparison with some Argoverse baseline methods. Our baseline (*) is represented by the system pipeline illustrated in Fig. \ref{fig:general_architecture_system_pipeline}, that is, LSTM based GAN with Multi-Head Self-Attention, without target points extractor. We conduct an ablation study to observe the influence of incorporating target points and class balance to our baseline. As expected, by explicitly defining the locations an agent is likely to be at a fixed prediction horizon for a given input trajectory and scene geometry, we are able to improve our baseline. Additionally, since nonlinear trajectories are more challenging than standard straight trajectories, we also observe how enforcing the class balance (straight, curve) during training is able to improve performance.

\begin{figure}[!ht]
    \centering
    \setlength{\tabcolsep}{2.0pt}
    \begin{tabular}{c}
    \includegraphics[width=0.50\linewidth]{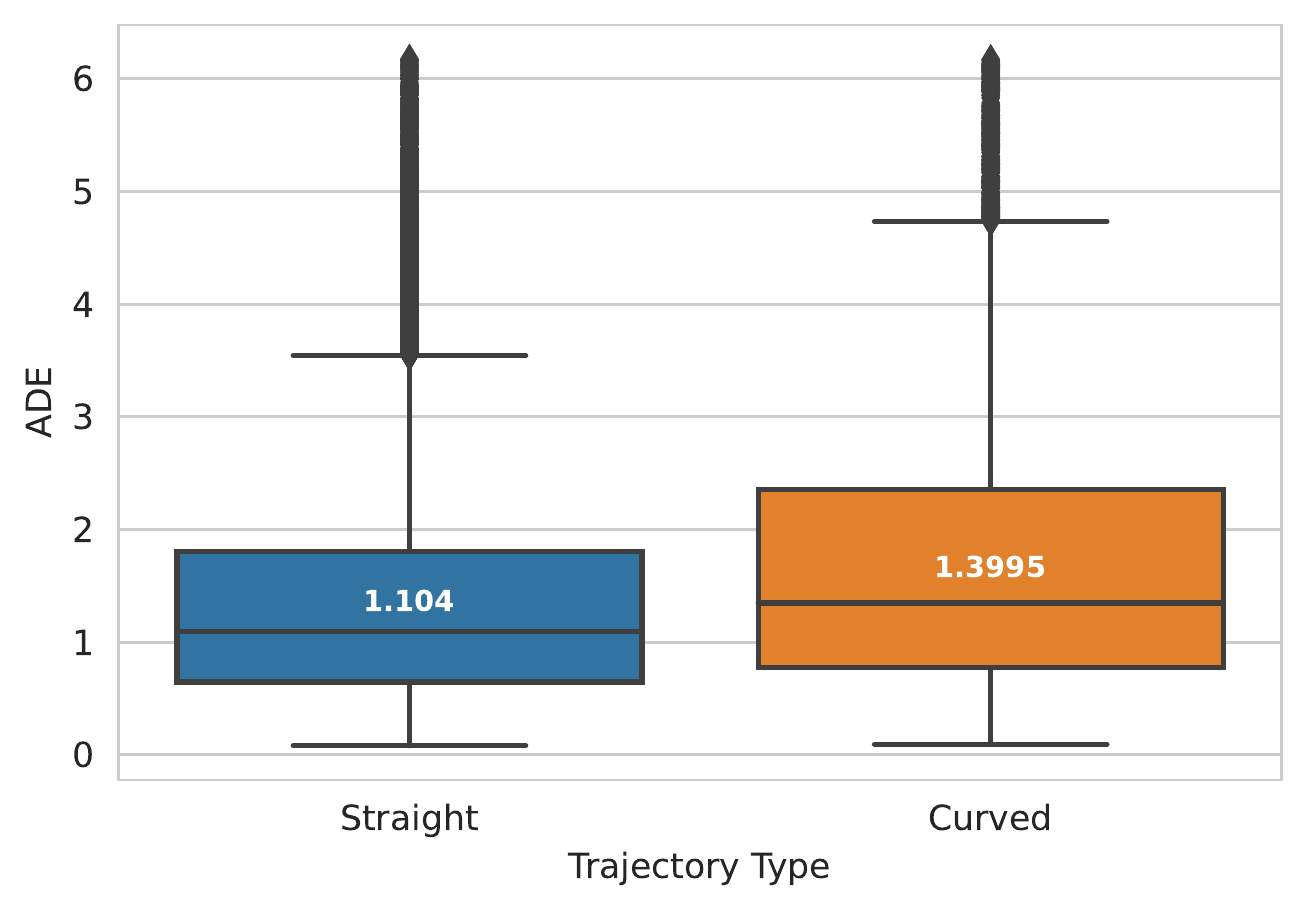} 
    \includegraphics[width=0.50\linewidth]{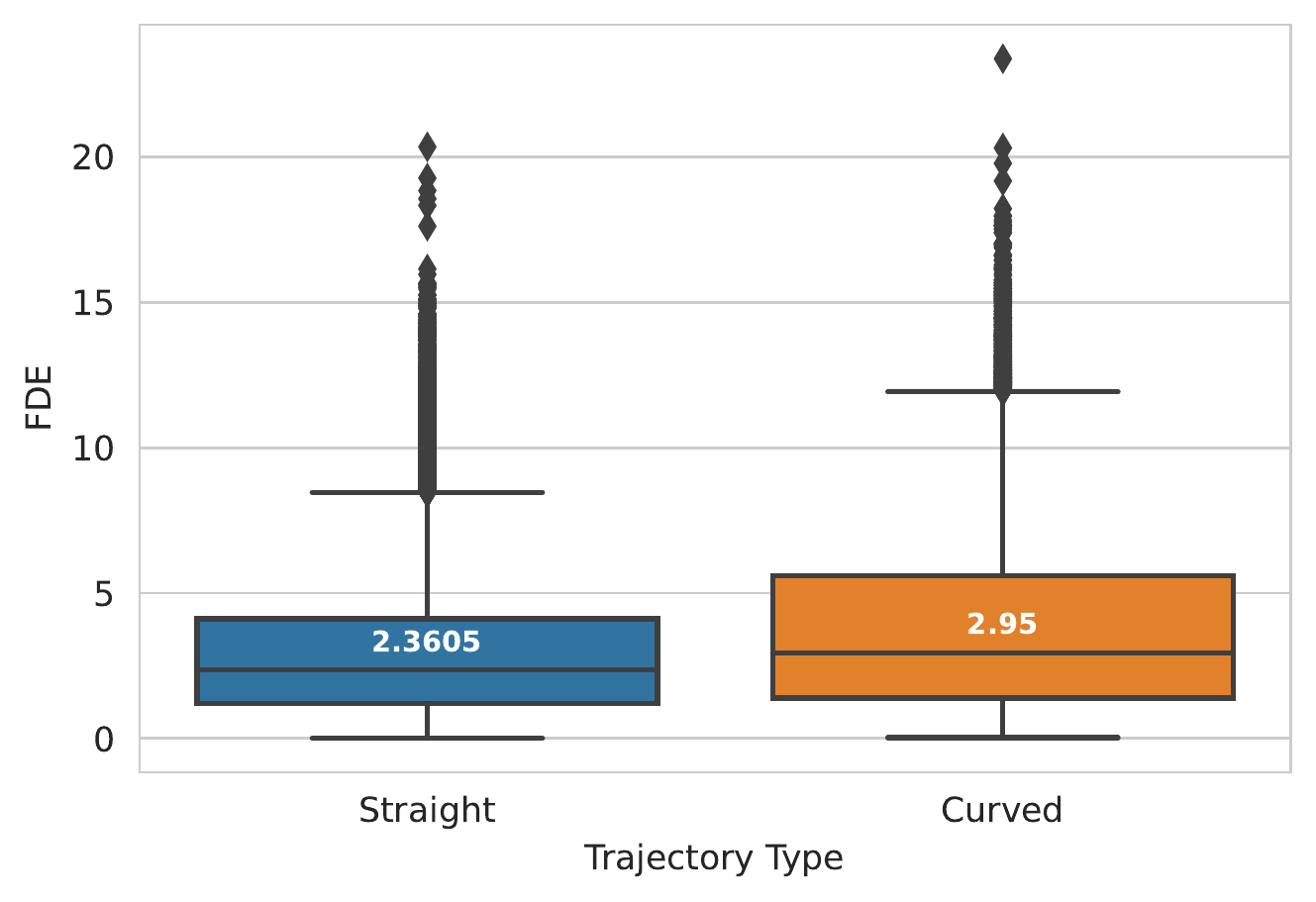}\tabularnewline
    \end{tabular}
    \caption{Statistical results on the Argoverse Validation Dataset. We show the boxplots for ADE and FDE metrics. We distinguish between straight and curved trajectories. We highlight the median (Q2) in each boxplot.}
    \label{fig:boxplots}
\end{figure}

\begin{figure*}[!ht]
    \centering
    \setlength{\tabcolsep}{2.0pt}
    \begin{tabular}{cccc}
    \fbox{\includegraphics[width=0.32\linewidth]{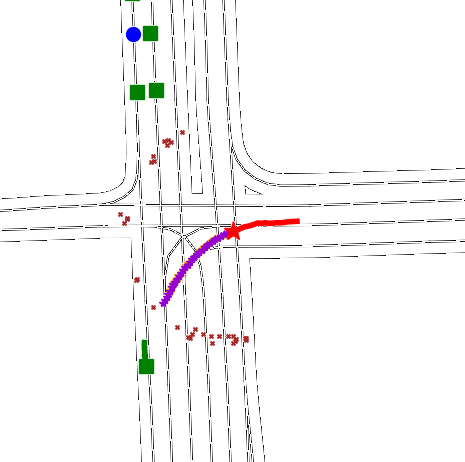}} &
    \fbox{\includegraphics[width=0.32\linewidth]{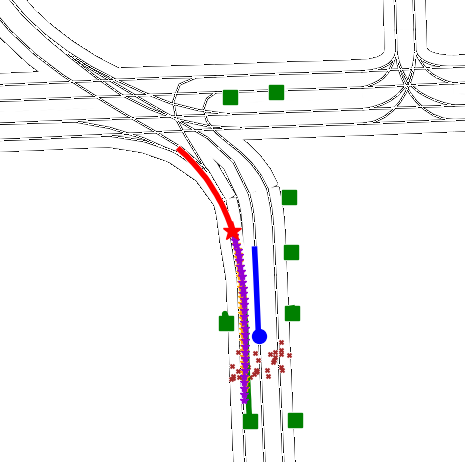}} &
    \fbox{\includegraphics[width=0.32\linewidth]{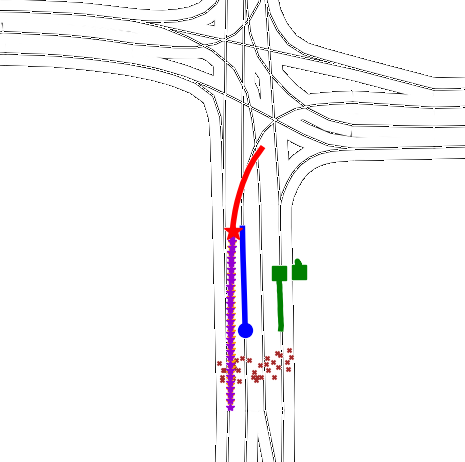}}
    \tabularnewline
    \fbox{\includegraphics[width=0.32\linewidth]{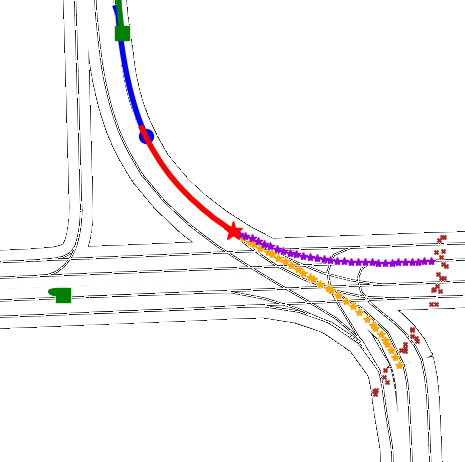}} & 
    \fbox{\includegraphics[width=0.32\linewidth]{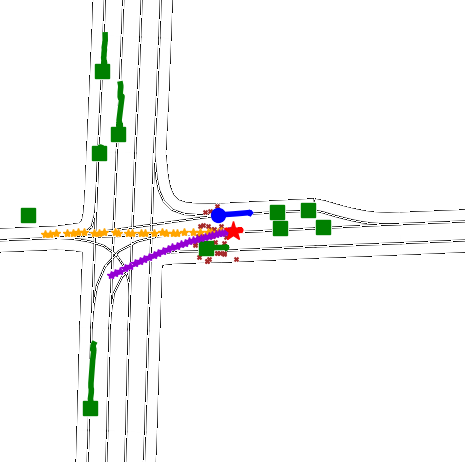}} &
    \fbox{\includegraphics[width=0.32\linewidth]{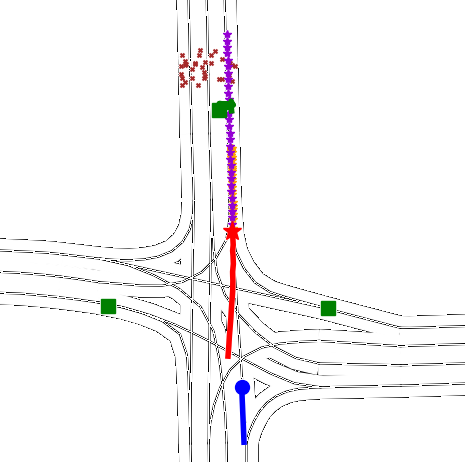}} \tabularnewline
    \end{tabular}
    \caption{Qualitative Results using our best model (including target points extraction and class balance). The legend is as follows: our vehicle (\textcolor{blue}{ego}), the target \textcolor{red}{agent}, and \textcolor{ForestGreen}{other agents}. We can also see the \textcolor{orange}{real} trajectory, the \textcolor{purple}{prediction} and potential \textcolor{brown}{goal-points}. Markers are current positions. First row shows feasible predicted trajectories, whilst the second one shows interesting challenging scenarios in which our current approach is not able to properly reason.}
    \label{fig:results1}
\end{figure*}

On the other hand, we analyze our performance on the Argoverse Validation Dataset, we use 31000 samples: 23012 straight and 7988 curved trajectories. We show in Figure~\ref{fig:boxplots} the boxplots for the ADE and FDE metrics. As stated before, our method, as most methods, struggles with curved trajectories, the overall ADE and FDE is "always" better for the straight trajectory cases. The median provides a robust estimator of our trajectories error. Note that we detected multiple outliers in our analysis, these are due to the unimodal nature of the predicted trajectories that makes difficult for the model to consider multiple possible hypotheses (multimodal). 

Fig. \ref{fig:results1} illustrates some qualitative results, all of them considering unimodal prediction towards the precomputed target points, meeting the physical and social constraints in the scenes. It can be clearly appreciated that a naive CTRV (Constant Turn Rate Velocity) could not generalize in these situations, where the vehicle can describe a curved future trajectory given a predominant straight input trajectory and viceversa. On top of that, second row illustrates quite interesting scenarios where the lack of reasoning and multimodality (producing a set of k-trajectories towards the set of target points) is revealed, being situations in which predicting accurately the end-point is difficult, and this the FDE is extremely high. Left image shows how we compute an unimodal prediction in the wrong direction of a split, even though target points are extracted very close to the groundtruth end point. Center image shows an extreme difficult situation, where the input trajectory is almost in the same place (probably the target agent was stopped in front a traffic light) whilst the groundtruth future trajectory is clearly an acceleration since the last observation frame. Finally, right image shows a deceleration because of the ahead obstacle whilst our model is not able to properly reason the presence of this obstacle in order to meet common sense safety constraints.

\section{CONCLUSIONS}
\label{section:conclusions_and_future_works}

Forecasting the future trajectories of surrounding actors in the scene is mandatory to achieve a safe planning, and thus, a crucial part of the Autonomous Driving stack. In this work we explore a LSTM Multi-Head Self Attention GANs for vehicle motion prediction using the Argoverse Motion Forecasting Benchmark 1.1. Our model considers both the physical and social context of the scene to predict the most plausible trajectory using a generative model, and achieves competitive results in comparison to other state-of-the-art methods regarding the case of unimodal prediction. In future works we plan to extend our model by incorporating stochastic multimodality and enriched attention over the physical context, specially focusing on the vector features of HDMap, implement an enhanced target-conditioned decoder to produce more feasible and realistic trajectories and run our model in the recently released Argoverse Motion Forecasting Benchmark 2.0, which allows for Multi-Agent evaluation.

\section*{ACKNOWLEDGMENT}
This work has been funded in part from the Spanish MICINN/FEDER through the Artificial Intelligence based modular Architecture Implementation and Validation for Autonomous Driving (AIVATAR) project (PID2021-126623OB-I00) and from the RoboCity2030-DIH-CM project (P2018/NMT- 4331), funded by Programas de actividades I+D (CAM), cofunded by EU Structural Funds and Scholarship for Introduction to Research activity by University of Alcalá.

Marcos Conde is with the Computer Vision Lab, University of Würzburg, supported by the Humboldt Foundation.

{
\bibliographystyle{ieeetr}
\bibliography{itsc_root.bib}
}

\end{document}